%% file: main.tex
\documentclass{article}

\PassOptionsToPackage{numbers, compress}{natbib}
 \usepackage[preprint]{neurips_2026}

\usepackage[utf8]{inputenc} 
\usepackage[T1]{fontenc}    
\usepackage{hyperref}       
\usepackage{url}            
\usepackage{booktabs}       
\usepackage{amsfonts}       
\usepackage{nicefrac}       
\usepackage{microtype}      
\usepackage{xcolor}         
\usepackage{bm}

\usepackage{amsthm}
\newtheorem{assumption}{Assumption}

\usepackage{multirow}
\usepackage{amsmath}   
\usepackage{graphicx}  
\usepackage{caption}
\usepackage{colortbl} 
\usepackage{algorithm}
\usepackage{algorithmic}

\newcommand{\E}{\mathbb{E}}
\newcommand{\KL}{\mathrm{KL}}
\newcommand{\Entropy}{\mathrm{H}}
\newcommand{\I}{\mathrm{I}}

\newcommand{\cand}{\mathcal{C}}

\usepackage{wrapfig}
\definecolor{lightgray}{rgb}{0.9, 0.9, 0.9}

\title{DepCap: Adaptive Block-Wise Parallel Decoding for Efficient Diffusion LM Inference}

\author{
Xiang Xia$^1$ \And
Wuyang Zhang$^1$\thanks{Corresponding Author} \And
Jiazheng Liu$^1$ \And
Cheng Yan$^1$ \And
Yanyong Zhang$^{1,2}$\and
$^1$University of Science and Technology of China \\
$^2$Institute of Artificial Intelligence, Hefei Comprehensive National Science Center \\
\texttt{xxia@stu.ecnu.edu.cn, \{liujiazheng25, yc\_sa22218099\}@mail.ustc.edu.cn,}\\
\texttt{\{wuyangz, yanyongz\}@ustc.edu.cn}
}

\begin{document}

\maketitle

\begin{abstract}


Diffusion language models (DLMs) have emerged as a promising alternative to autoregressive language generation due to their potential for parallel decoding and global refinement of the entire sequence. To unlock this potential, DLM inference must carefully balance generation quality and decoding speed. Recent block-wise DLM decoding methods improve this trade-off by performing diffusion-based decoding sequentially in blocks. However, existing methods typically rely on fixed block schedules or current-step local signals to determine block boundaries, and use conservative confidence-based parallel decoding to avoid conflicts, limiting the quality-speed trade-off. In this paper, we argue that block-wise DLM inference requires more suitable signals for its two core decisions: cross-step signals for determining block boundaries, and token-level conflict signals for parallel decoding. Based on this view, we propose DepCap, a training-free framework for efficient block-wise DLM inference. Specifically, DepCap instantiates the cross-step signal as the influence of the last decoded block and uses it to adaptively determine how far the next block should extend, while identifying a conflict-free subset of tokens for safe parallel decoding within each block, enabling substantial inference acceleration with negligible quality degradation. DepCap is a plug-and-play method applicable to various DLMs, and compatible with existing KV-cache strategies for block-wise DLM. An information-theoretic analysis further suggests that the cumulative last-block influence on a candidate block is approximately additive across tokens, supporting the proposed block-partitioning criterion. Experimental results show that DepCap achieves favorable speed-quality trade-offs across multiple DLM backbones and reasoning and coding benchmarks, with up to 5.63$\times$ speedup without significant performance degradation.

\end{abstract}

\section{Introduction}
\label{sec:introduction}
Diffusion language models (DLMs)~\citep{SEDD, DLMsuvery1,DLMsuvery2} have emerged as a promising alternative to autoregressive language generation because they can refine many token positions in parallel instead of decoding tokens strictly from left to right. Unlike autoregressive models, DLMs can use bidirectional attention during the generation process, allowing each position to be updated using both left and right context. This property makes DLMs appealing for flexible text generation, especially on tasks such as reasoning~\citep{d1,DCoLT} and coding~\citep{DiffuCoder}, where decoding efficiency and rich contextual modeling are both important. Recent open-source diffusion large language models (dLLMs), including the Mercury~\citep{Mercury}, Dream~\citep{Dream7b} and LLaDA~\citep{LLADA, LLADA2.0} families, further show that this paradigm is increasingly practical and promising. However, due to the multi-step denoising process and the need for global refinement during inference, DLMs inherently face a trade-off between generation quality and speed, making efficient inference a critical challenge~\citep{Fast-dLLM, dkv_cache}.

To fully unlock the parallel decoding potential of DLMs, recent work has increasingly adopted block-wise decoding strategies~\citep{LLADA, Fast-dLLM, Block_Diffusion} for inference. 
Instead of refining the entire unfinished sequence at every step, block-wise DLMs decode the output sequentially in blocks, which requires \textit{two key inference-time decisions: how to determine the next block boundary and which tokens to decode in parallel within the current block}. 
This strategy significantly reduces the refinement scope at each iteration while allowing each decoded block to serve as clean context for subsequent blocks, thereby improving the practicality of DLM inference. 
It also enables subsequent acceleration techniques, including KV-cache mechanisms~\citep{dkv_cache, Fast-dLLM, Sparse-dLLM} tailored to block-wise DLM inference.

However, existing block-wise methods still leave a substantial quality-speed gap. 
The main limitation is that the two key inference decisions in block-wise DLM decoding are still made from suboptimal signals. 
First, at the block level, most methods either use a fixed block size or determine the next boundary from signals available only at the current decoding step, rather than from signals that reflect the evolving multi-step denoising process of DLM inference.
Second, at the token level, many methods rely on conservative confidence-based parallel decoding strategies. Such strategies treat candidate positions largely independently and fail to account for token conflicts: lowering the confidence threshold may improve decoding speed, but it can also decode conflicting tokens simultaneously, leading to degraded quality~\citep{DAWN}. 
As a result, current methods do not fully exploit the signals needed for either adaptive block partitioning or within-block parallel decoding, leading to a suboptimal trade-off between generation quality and decoding speed.

\noindent\textbf{Problem.} How can block-wise DLM inference use more suitable signals for its two core decisions of block partitioning and within-block parallel decoding, so as to improve decoding speed while preserving generation quality and avoiding retraining?

\noindent\textbf{Contribution.} To address these challenges, we propose \textbf{DepCap}, a training-free dependency-aware framework for block-wise DLM inference. The central idea is to make the two inference decisions in block-wise decoding depend on more suitable signals. 
At the block level, DepCap goes beyond fixed schedules and current-step local cues by introducing a cross-step signal derived from the last decoded block.
Specifically, the proposed dependency-guided adaptive block partitioning strategy (DepGA-Block) measures how strongly the last decoded block influences future positions and combines this signal with predictive uncertainty through a dependency score to determine the next block boundary. 
Within each chosen block, a conflict-aware parallel decoding strategy (CAP-Decoding) explicitly detects token conflicts and decodes only a conflict-free subset in parallel, enabling more aggressive parallel decoding without sacrificing generation quality. 
DepCap is plug-and-play, compatible with existing block-wise cache strategies, and supported by an information-theoretic analysis showing that the cumulative influence of last decoded block is approximately additive across tokens, which helps explain the proposed block-partitioning criterion.

Experimental results verify that DepCap achieves favorable speed-quality trade-offs across different DLM backbones, including the LLaDA~\citep{LLADA, LLADA1.5} and the Dream~\citep{Dream7b} family, on reasoning and coding benchmarks. In LLADA family, DepCap attains an average 3.57$\times$ speedup over vanilla decoding strategy without significant performance degradation and in some cases improves both throughput and accuracy. For example, on the MBPP benchmark with LLaDA-1.5, DepCap achieves a 5.63$\times$ speedup while improving accuracy by 7.4\% relative.

The subsequent sections present the related work and preliminaries, describe the proposed DepCap method, show the empirical results, and conclude the paper.

\section{Related Work}
\label{sec:related-work}

\noindent\textbf{Diffusion language models.}
Diffusion language models (DLMs)~\citep{DLMsuvery1, DLMsuvery2} formulate text generation as an iterative denoising process rather than a strictly left-to-right prediction process. This paradigm originates from diffusion modeling for continuous domains~\citep{ddpm} and has been extended to discrete text generation~\citep{DiffuSeq, D3PM}. A key advantage of DLMs is their ability to perform parallel token updates and leverage bidirectional context during inference, enabling non-sequential generation that differs fundamentally from autoregressive generation. Recent work~\citep{diffugpt,LLADA,Dream7b, SDAR} has further scaled DLMs to stronger language modeling settings, showing that diffusion-based generation is becoming increasingly practical, with models reaching 100B parameters (e.g., LLaDA 2.0~\citep{LLADA2.0}). Compared to autoregressive language models, this non-sequential paradigm offers greater flexibility in generation but introduces the challenge of balancing generation quality and decoding speed.

\noindent\textbf{Block-wise decoding in DLMs.}
To improve the generation speed-quality trade-off of DLM inference, recent work adopts a block-wise decoding strategy~\citep{Block_Diffusion, Fast-dLLM, LLADA, fast-dLLMv2}, where the sequence is decoded sequentially in blocks rather than repeatedly refining unfinished positions at every step. Existing block-wise DLMs often rely on fixed block sizes, which ignores that the suitable block size can vary across inputs and even across different stages of the same generation process. Therefore, more recent approaches attempt to dynamically adjust block size. AdaBlock~\citep{AdaBlock} determines block boundaries by examining semantic structure, such as special tokens (e.g., ``\textbackslash n''), while Swordsman~\citep{Swordsman} partitions blocks based on entropy changes across adjacent positions. A concurrent work, GeoBlock~\citep{GeoBlock}, infers block size from attention-derived dependency geometry. However, these methods primarily rely on signals available at the current decoding step, neglecting the temporal dynamics inherent in the multi-step denoising process. As a result, they fail to exploit cross-step information that could provide stronger guidance for adaptive block size selection.

\noindent\textbf{Efficient inference for DLMs.}
To further unlock the parallel generation potential of DLMs, existing work mainly focuses on two directions: decoding strategies and cache techniques. 
For decoding strategies, recent methods accelerate inference via heuristic-based parallel decoding~\citep{Fast-dLLM, DAWN}, learned parallel decoding~\citep{LearningToParallel, dParallel}, denoising-process redesign~\citep{D2F}, and speculative decoding~\citep{FlashDLM,APD,DiffuSpec}.
Cache techniques~\citep{Fast-dLLM, dkv_cache, dLLM-Cache, d2cache} form the other direction, and their core idea is typically to refresh cached states at selected intervals based on the similarity of hidden states across adjacent decoding steps.
While these approaches show that substantial DLM inference acceleration is possible, achieving an effective balance between generation quality and decoding speed remains a significant challenge.

\section{Preliminaries}
\label{sec:preliminaries}

This section introduces diffusion language model inference and the block-wise decoding setting considered in this paper.

\noindent\textbf{Diffusion language model inference.}
Diffusion language models (DLMs) considered here adopt a masked language modeling formulation. Let $\mathcal{V}$ denote the token vocabulary, including the special token $\texttt{[MASK]}$. Given a prompt $x$, let
\(\bm{y} = (y_1,\dots,y_N) \in \mathcal{V}^N \)
denote the target sequence to be generated. In inference, the model takes as input the prompt $x$ followed by the current target sequence $\bm{y}$. At the beginning of inference, this sequence is initialized as 
\(\bm{y}^T = (\texttt{[MASK]},\dots,\texttt{[MASK]}).\)
At denoising step $t$, the current state of the target sequence is $\bm{y}^t = (y_1^t,\dots,y_N^t)\in \mathcal{V}^N$ . The masked positions at step $t$ are
\(
M^t = \{ i : y_i^t = \texttt{[MASK]} \}.
\)
At each step, the model first predicts token distributions for positions in $M^t$, and then a decoding strategy chooses a subset $S^t \subseteq M^t$ to decode. The decoding process can be defined as
\begin{equation}
y_i^{t-1} =
\begin{cases}
\hat{y}_i^t, & i \in S^t,\\
\texttt{[MASK]}, & i \in M^t \setminus S^t,\\
y_i^t, & i \notin M^t,
\end{cases}
\end{equation}
where $\hat{y}_i^t$ is the predicted token at position $i$. Repeating this iterative decoding process gradually removes mask tokens from $\bm{y}^t$ until all masked positions have been decoded, and the final sequence satisfies $\bm{y}^0 \in (\mathcal{V} \setminus \{\texttt{[MASK]}\})^N$. Unlike autoregressive decoding, each denoising step can update multiple positions in parallel while using bidirectional context.

\noindent\textbf{Block-wise decoding.}
To better balance generation quality and speed, recent DLM inference methods often partition the target sequence into a sequence of non-overlapping blocks
\(
\mathcal{B} = \{B_1,\dots,B_K\}.
\)
During inference, blocks are decoded sequentially: block $B_{k+1}$ starts only after block $B_k$ has been decoded.
For brevity, we define the current decoding block at step $t$ as $B^{t} \in \mathcal{B}$. Then block-wise decoding restricts the decoding subset to
\(
S^t \subseteq M^t \cap B^{t}.
\)
Equivalently, the sequence update becomes
\begin{equation}
y_i^{t-1} =
\begin{cases}
\hat{y}_i^t, & i \in S^t,\\
\texttt{[MASK]}, & i \in \left(M^t \cap B^{t}\right)\setminus S^t,\\
y_i^t, & i \notin M^t \cap B^{t},
\end{cases}
\end{equation}
which means that only positions in the current block are allowed to decode at step $t$.

\noindent\textbf{Two decisions in block-wise inference.}
Block-wise decoding introduces two inference decisions. The first is the \emph{block-boundary decision}: given the already decoded context, how large should the next block be? The second is the \emph{within-block decoding decision}: within the chosen block, which positions should be decoded together at the current denoising step? Existing block-wise DLM inference methods mainly differ in how they make these two decisions.

\section{The DepCap Method}
\label{sec:method}
This section presents DepCap, a training-free inference framework for block-wise diffusion language models (DLMs). We first overview DepCap, then describe the dependency-guided adaptive block partitioning strategy (DepGA-Block) and the conflict-aware parallel decoding strategy (CAP-Decoding), and finally give a compact analysis of the dependency score used for block partitioning.

\subsection{Overview}

\begin{figure*}[t]
  \centering
    \includegraphics[width=0.93\textwidth]{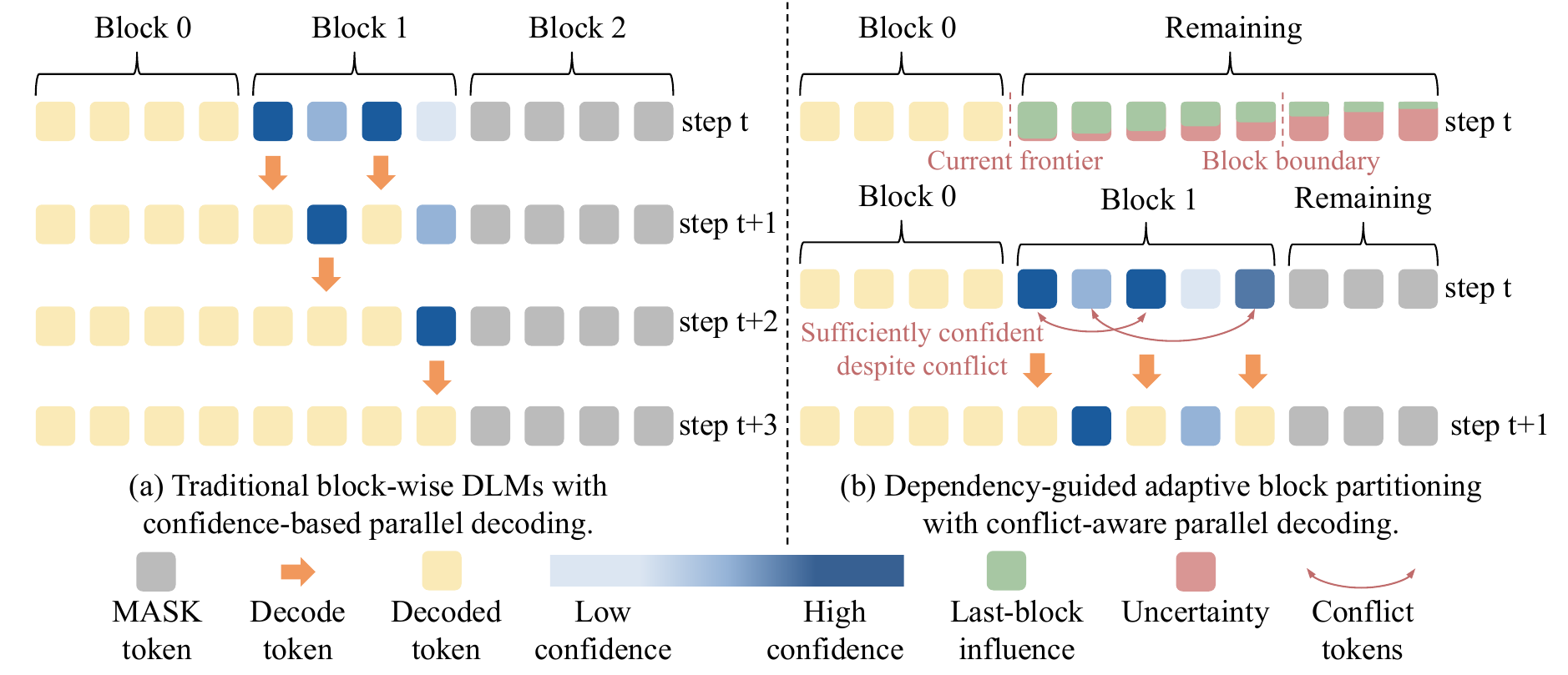}
  \caption{The framework of DepCap. (a) Traditional block-wise DLM inference uses a fixed block size and confidence-only parallel decoding. (b) DepCap adaptively chooses the next block boundary with DepGA-Block based on last-block influence and predictive uncertainty, and applies CAP-Decoding to safely decode a conflict-free subset of positions inside the current block. Darker color indicates higher confidence, and red bidirectional arrows denote token conflicts. In the DepGA-Block visualization, green and red indicate last-block influence and predictive uncertainty, respectively. A block boundary is chosen when predictive uncertainty becomes dominant over last-block influence.}
  \label{fig:DepCap}
  \vspace{-10pt}
\end{figure*}

DepCap makes two inference-time decisions at each stage of block-wise DLM decoding: how far the next block should extend, and which positions inside that block can be parallel decoded safely. For the first decision, DepGA-Block goes beyond fixed schedules and current-step local cues by using a cross-step signal derived from the last decoded block to guide block expansion. For the second, explicitly filters token conflicts that confidence-only decoding strategies may miss. Because both modules operate only on inference-time predictive signals, DepCap can be added on top of existing block-wise DLM inference processes without retraining, and is compatible with existing block-wise cache techniques. Figure~\ref{fig:DepCap} illustrates the overall DepCap framework. The following describes DepGA-Block and CAP-Decoding modules in detail.

\subsection{Dependency-Guided Adaptive Block Partitioning Strategy}

In order to adapt block boundaries to the evolving DLM generation process, we propose the dependency-guided adaptive block partitioning strategy (DepGA-Block). Existing methods typically determine block boundaries either with a fixed block size~\citep{Fast-dLLM} or from signals available at the current decoding step, such as semantic structure~\citep{AdaBlock}, entropy~\citep{Swordsman}, or attention~\citep{GeoBlock}. In contrast, DepGA-Block follows the multi-step denoising behavior of DLM inference and introduces a cross-step signal derived from the last decoded block to guide the next block boundary.

Specifically, suppose the last decoded block $B$ has just been added to the decoded context. For each candidate future position $k$, let $p_k^{\text{prev}}$ denote the predictive distribution before decoding $B$, and let $p_k^{\text{curr}}$ denote the predictive distribution after decoding $B$. DepGA-Block measures the influence of the last decoded block by the KL divergence between these two predictive distributions:
\begin{equation}
  I_k = \KL\!\left(p_k^{\text{curr}} \,\middle\|\, p_k^{\text{prev}}\right),
\end{equation}
which we call \emph{last-block influence}. $I_k$ measures how much decoding the last block shifts the marginal predictive distribution at position $k$ and a larger $I_k$ means that the last decoded block provides strong cross-step information for deciding the next block boundary.
To complement this influence signal, DepGA-Block measures predictive uncertainty by the Shannon entropy of the predictive distribution: $H_k = \Entropy\!\left(p_k^{\text{curr}}\right).$
Large entropy indicates that the position remains uncertain even after the last block has been decoded.
Together, $I_k$ and $H_k$ capture whether future positions are still strongly supported by the latest decoded context or are already dominated by uncertainty.

In practice, both signals are smoothed and normalized within a local candidate window of size $L_\text{window} = \min(L_{\max}, L_\text{remain})$, where $L_\text{remain}$ denotes the number of remaining masked tokens after the current decoding frontier, This yields the normalized signals $\tilde I_k$ and $\tilde H_k$. DepGA-Block then defines the \emph{dependency score} to determine the block size:
\begin{equation}
  S_k = \tilde I_k - \lambda \tilde H_k,
  \label{eq:score}
\end{equation}
where $\lambda > 0$ controls the uncertainty penalty, so that the first position where $S_k < 0$ marks the point at which uncertainty outweighs the support provided by the last decoded block, with larger $\lambda$ giving more conservative block expansion. Starting from the current frontier, DepCap scans forward and finds the first position where the score drops below zero. The next block ends immediately before that position, and the final block length $L$ is clipped to the range $[L_{\min}, L_\text{window}]$. For the initial step, the first block uses $L = L_{\min}$ as a conservative cold start. An information-theoretic interpretation of this dependency score is given in Section~\ref{sec:analysis}.

DepGA-Block expands the next block when the last decoded block continues to provide strong support for nearby future positions, and contracts it when uncertainty begins to dominate. Importantly, all signals used here are available from the model's predictive distributions during inference.

\subsection{Conflict-Aware Parallel Decoding Strategy}

Traditional confidence-based decoding strategies often employ conservative thresholds to preserve generation quality, but this inevitably limits the parallel generation capability of DLMs. To enhance within-block parallelism without noticeably hurting generation quality, we introduce the conflict-aware parallel decoding strategy (CAP-Decoding), which explicitly models token conflicts and constructs a safe token subset for parallel decoding, as summarized in Algorithm~\ref{alg:cap-decoding}.

Specifically, once the next block $B^{t}$ has been chosen, CAP-Decoding decides which masked positions within that block can be decoded simultaneously. Given the token confidence $c_i = \max_v p_i(v)$ at position $i$, we define the candidate pool in the current block as: 
\begin{equation}
  \cand = \{i : i \in M^t \cap B^{t},\; c_i \ge \tau_{\text{low}}\}
\end{equation}
where $M^t$ denotes the set of still masked tokens in the sequence. While confidence aids in filtering potential candidates, it does not solely guarantee that a set of positions can be safely parallel decoded.

To further determine which candidates can be decoded together, CAP-Decoding employs a lightweight score to quantify token conflicts between candidate positions. For any pair of positions $(i,j) \in \cand \times \cand$, the score is computed as:
\begin{equation}
  D_{ij} = \log p_i(\hat{y}_j) + \log p_j(\hat{y}_i),
  \label{eq:conflict}
\end{equation}
where $\hat{y}_i$ and $\hat{y}_j$ are the current token predictions at positions $i$ and $j$, respectively. 
A high $D_{ij}$ score signifies a strong conflict, where the probability distributions of the two positions are heavily entangled under the current distributions.
To avoid committing unsafe tokens simultaneously, positions are flagged as conflicting when $D_{ij} > \gamma$ and are thus excluded from being decoded in parallel.

\begin{algorithm}[t]
\caption{Conflict-Aware Parallel Decoding Strategy (CAP-Decoding)}
\label{alg:cap-decoding}
    \begin{algorithmic}[1]
    \REQUIRE Current block $B^{t}$, masked-position set $M^t$, predictive distributions $\{p_i\}_{i \in B^{t}}$, confidence thresholds $\tau_{\text{low}}, \tau_{\text{high}}$, and conflict threshold $\gamma$
    \ENSURE 
    \STATE Compute token confidence $c_i = \max_v p_i(v)$ for each candidate $i \in M^t \cap B^{t}$
    \STATE Build candidate pool $\cand = \{i : i \in M^t \cap B^{t},\; c_i \ge \tau_{\text{low}}\}$
    \FOR{each $(i,j) \in \cand \times \cand$ with $i \neq j$}
        \STATE Let $\hat{y}_i = \arg\max_v p_i(v)$ and $\hat{y}_j = \arg\max_v p_j(v)$
        \STATE Compute conflict score $D_{ij} = \log p_i(\hat{y}_j) + \log p_j(\hat{y}_i)$
    \ENDFOR
    \STATE Initialize $\mathcal{S} = \{i \in \cand : c_i \ge \tau_{\text{high}}\}$  \hfill \texttt{$\triangleright$} Phase 1: High-confidence priority
    \STATE Remove $\mathcal{S}$ and all candidates that conflict with any candidate in $\mathcal{S}$ from $\cand$ 
    \STATE Sort the remaining candidates $\cand$ in descending order of $c_i$ \hfill \texttt{$\triangleright$} Phase 2: Greedy completion
    \FOR{each candidate $i$ in sorted order}
        \STATE Add candidate $i$ to $\mathcal{S}$
        \STATE Remove candidate $i$ and the candidates that conflict with $i$ from $\cand$ 
    \ENDFOR
    \IF{$\mathcal{S} = \emptyset$}
    \STATE Add the candidate $i = \arg\max_{i \in M^t \cap B^t} c_i$ to $\mathcal{S}$
    \ENDIF
    \RETURN Safe subset $\mathcal{S} \subseteq M^t \cap B^{t}$ for parallel decoding
    \end{algorithmic}
\end{algorithm}

Then, CAP-Decoding selects the parallel subset in two phases. In the high-confidence priority phase, CAP-Decoding initializes the safe subset $\mathcal{S}$ with candidates satisfying $c_i \ge \tau_{\text{high}}$, and removes from the pool $\cand$ both these selected candidates and the remaining candidates that conflict with them. In the greedy completion phase, CAP-Decoding sorts the remaining candidates in $\cand$ by descending confidence, iteratively adds the top candidate $i$ to $\mathcal{S}$, and removes from $\cand$ both the candidate $i$ and the candidates that conflict with it. As a safeguard, CAP-Decoding always decodes the highest-confidence token in the current block.

This procedure allows CAP-Decoding to decode more positions in parallel than a conservative confidence-only decoding strategy, thereby avoiding the significant quality degradation caused by simultaneously decoding conflicting tokens.

\subsection{Analysis of the Dependency Score}
\label{sec:analysis}

This subsection gives a compact information-theoretic interpretation of the dependency score to explain the form of the block partitioning strategy in DepGA-Block.

Let $c$ denote the decoded context before the last decoded block, let $B \sim p(B \mid c)$ denote that block, and let $Z_{1:L} = (Z_1,\dots,Z_L)$ denote a candidate next block. For a realized $B=b$, the online quantity used by DepCap is
\begin{equation}
  I_k(b)
  =
  \KL\!\left(
    p(Z_k \mid c, b)
    \,\middle\|\,
    p(Z_k \mid c)
  \right).
\end{equation}
Its expectation is
\begin{equation}
  I_k^{\mathrm{exp}}
  =
  \E_{B \sim p(\cdot \mid c)}[I_k(B)]
  =
  \I(Z_k; B \mid c).
\end{equation}
Thus the last-block influence used by DepCap is the realized counterpart of a conditional mutual-information term.

\begin{assumption}[Local overlap]
\label{ass:local-overlap}
For the short frontier-local candidate blocks considered by DepCap, the information that the last decoded block provides to different future positions has limited overlap.
\end{assumption}

Under the Assumption\ref{ass:local-overlap}, the cumulative influence of the last decoded block on a candidate block,
\(
  \mathcal{I}(L) = \I(Z_{1:L}; B \mid c),
\)
is approximately additive across positions:
\begin{equation}
  \mathcal{I}(L)
  =
  \sum_{k=1}^{L} I_k^{\mathrm{exp}} - \varepsilon(L)
  \approx
  \sum_{k=1}^{L} I_k^{\mathrm{exp}},
\end{equation}
where $\varepsilon(L)$ is an overlap correction term. The Appendix~\ref{app:Analysis} gives the detailed derivation.

To account for predictive ambiguity, DepCap also considers
$H_k = \Entropy(Z_k \mid c, B).$
A natural design objective is that the cumulative support from the last decoded block should dominate the cumulative uncertainty of the candidate positions:
\begin{equation}
  \mathcal{I}(L)
  \geq
  \lambda \sum_{k=1}^{L} \E[H_k].
\end{equation}

Replacing $\mathcal{I}(L)$ by its additive approximation motivates a per-position criterion of the form $I_k - \lambda H_k$, which is exactly the structure of the dependency score in Equation~\eqref{eq:score}. In implementation, DepCap computes this score from the predictive distributions immediately before and after the last block is decoded, which provides a tractable approximation to the idealized quantities above.

\section{Experiments}
\label{sec:experiments}

To evaluate whether the proposed DepCap improves the efficiency of block-wise DLM inference while preserving generation quality, we conduct extensive experiments on multiple representative DLM backbones and benchmarks. This section first introduces the experimental settings, and then presents the overall results, ablation study, and hyperparameter analysis. The experiments are structured to address the following key questions:
\begin{enumerate}
\setlength{\itemsep}{0pt}
  \item[\textbf{Q1:}] Does DepCap improve the overall speed-quality trade-off of block-wise DLM inference?
  \item[\textbf{Q2:}] Does DepCap remain effective when combined with existing block-wise cache techniques?
  \item[\textbf{Q3:}] How much do DepGA-Block and CAP-Decoding each contribute to the final gains?
  \item[\textbf{Q4:}] How does DepCap behave under different hyperparameter choices?
\end{enumerate}
These questions are answered sequentially in the following subsections. Our code is available at \url{https://github.com/X-Xia0828/DepCap}.

\subsection{Experimental Settings}

We conduct comprehensive evaluations on three representative DLM backbones, specifically LLaDA-8B-Instruct~\citep{LLADA}, Dream-v0-base-7B~\citep{Dream7b}, and LLaDA-1.5~\citep{LLADA1.5}, to validate the effectiveness of the proposed DepCap. Additionally, We compare DepCap against the established block-wise DLM framework, Fast-dLLM~\citep{Fast-dLLM} and the adaptive block partition method AdaBlock~\citep{AdaBlock}.

\noindent\textbf{Hyperparameter settings.}
Unless otherwise specified, all methods use a generation length of $L_{\mathrm{gen}} = 256$. In DepCap, DepGA-Block uses $\lambda = 1.2$, $L_{\min} = 8$, and $L_{\max} = 128$, while CAP-Decoding uses $\tau_{\text{low}} = 0.8$, $\tau_{\text{high}} = 0.95$, and $\gamma = -16.0$. For the confidence-based decoding strategy, we use a threshold of $0.9$ with fixed block sizes $|B_0| \in \{16, 32, 64\}$. For AdaBlock, we use its default hyperparameter setting, with ``\textbackslash n'' as the detected token and default block size $B_{0} = 32$.

\noindent\textbf{Benchmarks and metrics.}
We evaluate all methods on four benchmarks: GSM8K (5-shot)~\citep{GSM8K} and Math-500 (4-shot)~\citep{math500} for mathematical and reasoning generation, and MBPP (3-shot)~\citep{mbpp} and HumanEval (0-shot)~\citep{humaneval} for code generation. We report task accuracy (Acc), tokens per second (TPS), and the number of function evaluations (NFE), where TPS measures throughput and NFE reflects computational cost during decoding. Among these metrics, TPS is computed over the full generated sequence until the end-of-sequence token is reached. Higher task accuracy and TPS indicate better generation quality and faster decoding, respectively, while lower NFE signifies fewer denoising steps. All experiments are conducted on a single NVIDIA A100 40GB GPU, and all benchmark evaluations are run with the \texttt{lm-eval} library to ensure a fair comparison. 

\subsection{Main Results}

\begin{table*}[t]
  \centering
  \caption{Main results on three DLM backbones and four benchmarks. We report task accuracy (Acc), throughput in tokens per second (TPS), and number of function evaluations (NFE) under different block partitioning strategy and decoding strategies. ``Vanilla" denotes decoding only the highest confidence token within the current block at each step and ``Confidence'' denotes standard confidence-based parallel decoding. Best results are \textbf{bolded}, and second-best are \underline{underlined}.}
  \resizebox{1.0\textwidth}{!}{
    \begin{tabular}{c|c|ccc|ccc|ccc|ccc}
    \toprule
    \multirow{2}{*}{\textbf{Parallel Decode}} & 
    \multirow{2}{*}{\textbf{Block}} & 
    \multicolumn{3}{c|}{\textbf{GSM8K (5-shot)}} & 
    \multicolumn{3}{c|}{\textbf{Math-500 (4-shot)}} & 
    \multicolumn{3}{c|}{\textbf{MBPP (3-shot)}} & 
    \multicolumn{3}{c}{\textbf{HumanEval (0-shot)}} \\
    \cmidrule(lr){3-5} \cmidrule(lr){6-8} \cmidrule(lr){9-11} \cmidrule(lr){12-14}
     &  
     & TPS $\Uparrow$ & NFE $\Downarrow$ & Acc $\Uparrow$ 
     & TPS $\Uparrow$ & NFE $\Downarrow$ & Acc $\Uparrow$ 
     & TPS $\Uparrow$ & NFE $\Downarrow$ & Acc $\Uparrow$ 
     & TPS $\Uparrow$ & NFE $\Downarrow$ & Acc $\Uparrow$ \\
    \midrule
    \rowcolor{lightgray} 
    \multicolumn{14}{c}{\textbf{LLaDA-8B-Instruct}} \\ 
    \midrule
    Vanilla & $|B_0|=32$ & 6.0$_{\times 1.00}$ & 256.0 & 77.6 & 7.9$_{\times 1.00}$ & 256.0 & 39.2 & 5.4$_{\times 1.00}$ & 256.0 & 40.8 & 16.7$_{\times 1.00}$ & 256.0 & \underline{42.7}\\
    \midrule
    \multirow{5}{*}{Confidence} 
      & $|B_0|=16$ & 18.3$_{\times 3.05}$ & 84.1 & 78.3 & 19.7$_{\times 2.49}$ & 104.4 & \underline{40.2} & 18.3$_{\times 3.39}$ & 75.9 & \textbf{41.2} & 51.0$_{\times 3.05}$ & 86.4 & \textbf{43.3} \\
      
      & $|B_0|=32$ & 19.6$_{\times 3.27}$ & 78.6 & 77.3 & 20.7$_{\times 2.62}$ & 99.1  & 39.6 & 19.7$_{\times 3.65}$ & 70.1 & \underline{41.0} & 56.5$_{\times 3.38}$ & 77.5 & \textbf{43.3} \\
      
      & $|B_0|=64$ & 20.1$_{\times 3.35}$ & 75.6 & 77.5 & 21.1$_{\times 2.67}$ & 97.0 & 37.8 & 19.9$_{\times 3.69}$ & \underline{68.2} & 40.8 & 57.2$_{\times 3.43}$ & 76.4 & 41.5 \\

      & AdaBlock & 18.4$_{\times 3.07}$ & 83.9 & 77.9 & 19.8$_{\times 2.51}$ & 103.5 & 40.0 & 19.4$_{\times 3.59}$ & 76.2 & 40.4 & 50.4$_{\times 3.02}$ & 86.2 & 40.9 \\

      & DepGA-Block & 18.7$_{\times 3.12}$ & 81.4 & \textbf{79.2} & 19.9$_{\times 2.52}$ & 101.8 & \textbf{40.4} & 18.8$_{\times 3.48}$ & 73.6 & \underline{41.0} & 53.4$_{\times 3.20}$ & 80.6 & \underline{42.7} \\
    \midrule
    \multirow{2}{*}{CAP-Decoding} 
      & $|B_0|=32$ & \textbf{21.6}$_{\times 3.60}$ & \textbf{71.5} & 77.3 & \textbf{22.8}$_{\times 2.89}$ & \textbf{90.9} & 39.2 & \textbf{21.2}$_{\times 3.93}$ & \textbf{65.7} & 40.6 & \textbf{60.4}$_{\times 3.62}$ & \textbf{72.0} & \textbf{43.3} \\
      
      & DepGA-Block & \underline{21.4}$_{\times 3.57}$ & \underline{74.8} & \underline{78.8} & \underline{22.2}$_{\times 2.81}$ & \underline{94.8} & 39.8 & \underline{20.8}$_{\times 3.85}$ & 68.6 & 40.8 & \underline{59.8}$_{\times 3.58}$ & \underline{74.8} & \underline{42.7} \\

      \midrule
    \rowcolor{lightgray} 
    \multicolumn{14}{c}{\textbf{Dream-v0-base-7B}} \\ 
    \midrule
    Vanilla & $|B_0|=32$ & 7.8$_{\times 1.00}$
    & 256.0 & \textbf{75.1}  & 9.7$_{\times 1.00}$  &  256.0 & \textbf{42.2}  & 10.2$_{\times 1.00}$  & 256.0 & 50.8   & 20.7$_{\times 1.00}$  & 256.0  & 49.4 \\
    \midrule
    \multirow{5}{*}{Confidence} 
      & $|B_0|=16$ & \textbf{13.7}$_{\times 1.76}$  & \underline{148.1}  & 74.7  & 23.7$_{\times 2.44}$  & 105.5  & 40.8  & 27.7$_{\times 2.72}$  & 94.1 & \textbf{53.2}   & 38.8$_{\times 1.87}$  & 137.0 & \textbf{55.5}  \\
      
      & $|B_0|=32$ & 13.2$_{\times 1.69}$  & 152.4  & 74.5  & 24.4$_{\times 2.52}$  & 102.3 & 40.4 & 28.6$_{\times 2.80}$  & 90.8  & 51.8  & 39.0$_{\times 1.88}$  & 135.3  & 53.0 \\
      
      & $|B_0|=64$ & 12.8$_{\times 1.64}$  & 157.9  & 74.6 & \textbf{24.9}$_{\times 2.57}$  & 99.5  & \underline{41.0}  & 28.9$_{\times 2.83}$ & 89.1  & 52.2   & 39.4$_{\times 1.90}$ & 131.8  & \underline{53.7} \\

      & AdaBlock & 13.1$_{\times 1.68}$ & 154.9  & \underline{75.1}  & 24.7$_{\times 2.55}$ & 101.7 & 39.8 & 28.2$_{\times 2.76}$ & 93.7 & 52.6 & 38.6$_{\times 1.86}$ & 137.9 & \underline{53.7}  \\

      & DepGA-Block & \underline{13.5}$_{\times 1.73}$ & 149.0  & 75.0 & 24.0$_{\times 2.47}$ & 103.2  & \underline{41.0} & 28.7$_{\times 2.81}$ & 90.1  & 52.6  & 39.5$_{\times 1.91}$ & 132.5  & 50.6  \\
    \midrule
    \multirow{2}{*}{CAP-Decoding} 
      & $|B_0|=32$ & 12.7$_{\times 1.63}$ & 149.9  & 74.1  & \underline{24.7}$_{\times 2.55}$ & \textbf{95.4}  & 40.4  & \underline{29.0}$_{\times 2.84}$ & \underline{85.2}  & 52.4  & \underline{39.8}$_{\times 1.92}$ & \underline{127.3}  & 50.6  \\
      
      & DepGA-Block & 13.0$_{\times 1.67}$ & \textbf{146.6}  & 74.8  & 24.4$_{\times 2.52}$ & \underline{97.2}  & 40.8  & \textbf{29.2}$_{\times 2.86}$ & \textbf{84.6}  & \underline{52.8}  & \textbf{40.6}$_{\times 1.96}$ & \textbf{125.3}   & 50.6  \\

      \midrule
    \rowcolor{lightgray} 
    \multicolumn{14}{c}{\textbf{LLaDA-1.5}} \\ 
    \midrule
    Vanilla & $|B_0|=32$ & 5.6$_{\times 1.00}$ & 256.0 & 81.1  & 7.1$_{\times 1.00}$ & 256.0  & 39.8  & 1.9$_{\times 1.00}$ & 256.0 & 37.8  & 5.9$_{\times 1.00}$ & 256.0  & \underline{41.5} \\
    \midrule
    \multirow{5}{*}{Confidence} 
      & $|B_0|=16$ & 17.5$_{\times 3.12}$ & 81.9  & 81.4  & 17.8$_{\times 2.51}$ & 101.0  & 40.0  & 10.4$_{\times 5.47}$ & 48.7  & \underline{40.2} & 14.2$_{\times 2.41}$ & 106.9  & 40.9 \\
      
      & $|B_0|=32$ & 18.9$_{\times 3.38}$ & 75.5  & 81.2  & 18.9$_{\times 2.66}$ & 95.2  & 39.4 & \underline{11.6}$_{\times 6.11}$ & 42.1 & 37.6 & 15.3$_{\times 2.59}$ & 99.6  & 40.2 \\
      
      & $|B_0|=64$ & 19.5$_{\times 3.48}$ & \underline{72.8}  & 80.7  & 19.3$_{\times 2.72}$ & \underline{91.7}  & 38.0 & 9.2$_{\times 4.84}$ & \textbf{40.1}  & 25.4 & 16.1$_{\times 2.73}$ & \underline{93.4}  & \textbf{42.1} \\

      & AdaBlock & 17.8$_{\times 3.18}$ & 80.1  & \underline{82.1}  & 18.5$_{\times 2.61}$ & 97.6  & 38.8 & 9.8$_{\times 5.16}$ & 51.8 & 36.4 & 14.1$_{\times 2.39}$ & 110.9  & 37.8 \\

      & DepGA-Block & 18.1$_{\times 3.23}$ & 80.1  & 80.6  & 18.1$_{\times 2.55}$ & 100.0  & \underline{41.6} & 10.3$_{\times 5.42}$ & 50.1  & 40.0  & 14.8$_{\times 2.51}$ & 105.6  & 39.6 \\
    \midrule
    \multirow{2}{*}{CAP-Decoding} 
      & $|B_0|=32$ & \textbf{20.6}$_{\times 3.68}$ & \textbf{69.1} & \underline{82.1} & \textbf{20.2}$_{\times 2.85}$ & \textbf{88.8} & 40.0 & \textbf{12.2}$_{\times 6.42}$ & \underline{40.6} & 39.0 & \textbf{16.4}$_{\times 2.78}$ & \textbf{93.0} & \textbf{42.1} \\
      
      & DepGA-Block & \underline{20.1}$_{\times 3.59}$ & 73.9 & \textbf{82.2} & \underline{19.7}$_{\times 2.77}$ & 94.1 & \textbf{42.2} & 10.7$_{\times 5.63}$ & 48.9 & \textbf{40.6} & \underline{16.2}$_{\times 2.75}$ & 98.5 & \underline{41.5} \\
    \bottomrule
    \end{tabular}
  }
  \label{tab:reorganized_parallel_decode}
  \vspace{-10pt}
\end{table*}

\noindent\textbf{To answer Q1.} We evaluate DepCap on three DLM backbones, namely LLaDA-8B-Instruct, Dream-v0-base-7B, and LLaDA-1.5, across four benchmarks: GSM8K, Math-500, MBPP, and HumanEval. The main results are reported in Table~\ref{tab:reorganized_parallel_decode}. \textbf{\textit{A first observation is that no single fixed block size dominates across models and tasks}}. Smaller blocks are often safer but leave substantial parallelism unused, whereas larger blocks can improve throughput at the cost of noticeable quality degradation. This directly motivates adaptive block partitioning and DepGA-Block indeed provides a more favorable speed-quality trade-off than rigid fixed schedules in most settings.

Compared with AdaBlock, DepGA-Block is generally more stable on the LLaDA family. While AdaBlock can occasionally match the best accuracy, DepGA-Block more often achieves a better balance between accuracy, throughput, and NFE. This suggests that using the influence of the last decoded block provides a more reliable signal for block partitioning than heuristics based only on the information available at the current decoding step.

When combined with CAP-Decoding, the full DepCap method achieves the strongest overall speed-quality trade-off on the LLaDA family. On MBPP with LLaDA-1.5, for instance, the full method achieves 5.63$\times$ speedup while improving accuracy by 7.4\% relative. More broadly, averaged over the eight LLaDA task-model pairs in the Table~\ref{tab:reorganized_parallel_decode}, the full DepCap framework achieves a 3.57$\times$ speedup compared with vanilla LLaDA decoding without significant performance degradation. These results indicate that dependency-guided adaptive block partitioning and conflict-aware within-block decoding work best when combined to achieve a better balance between generation quality and decoding speed, which answers Q1.

\begin{wrapfigure}{r}{0.53\textwidth}
  \centering
    \vspace{-10pt}
    \includegraphics[width=0.53\textwidth]{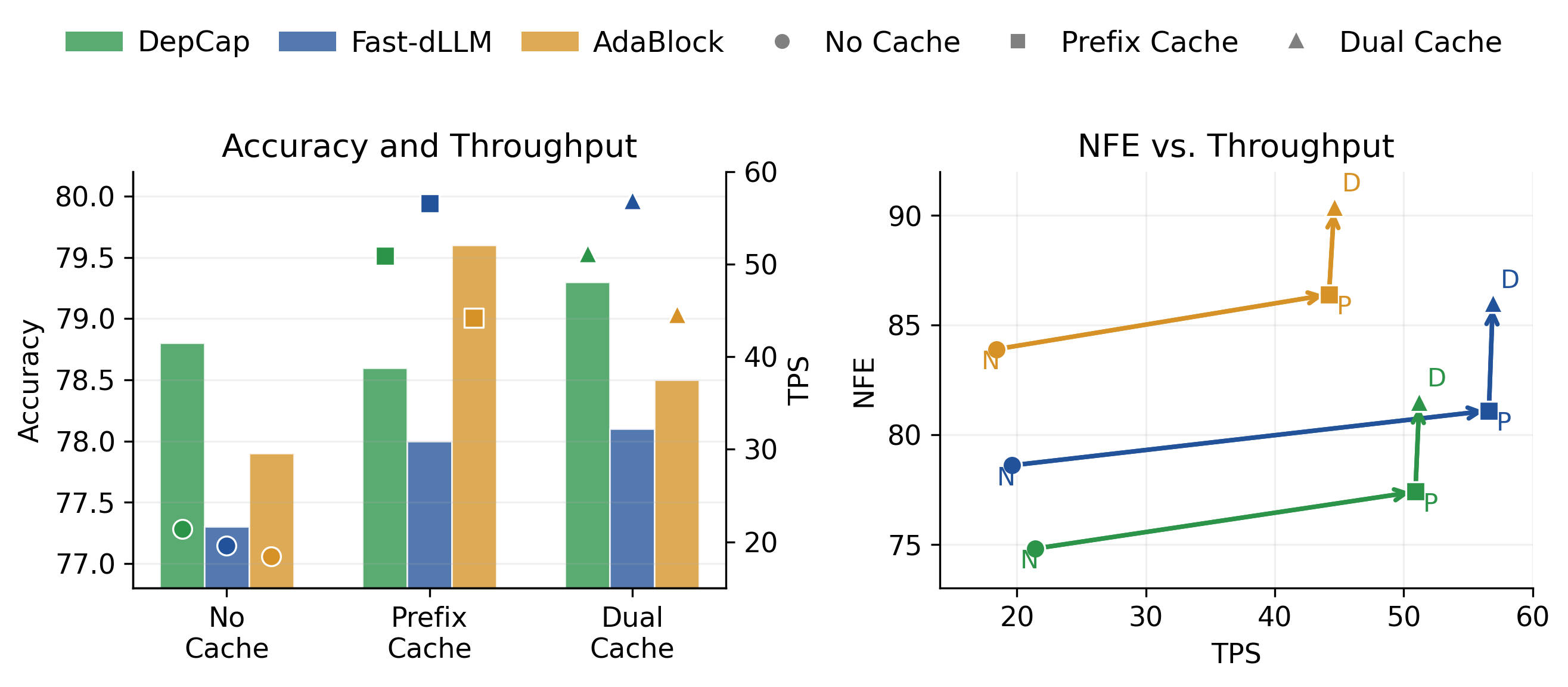}
  \caption{Cache-compatible results under No Cache, Prefix Cache, and Dual Cache. Left: accuracy versus throughput. Right: NFE versus throughput.}
  \label{fig:q2-cache}
  \vspace{-15pt}
\end{wrapfigure}

\noindent\textbf{To answer Q2.} We further evaluate whether DepCap remains effective when combined with existing block-wise cache techniques on LLaDA-8B-Instruct over the GSM8K benchmark. We compare DepCap with Fast-dLLM~\citep{Fast-dLLM} and AdaBlock~\citep{AdaBlock} under three cache settings: No Cache, Prefix Cache~\citep{Fast-dLLM} and Dual Cache~\citep{Fast-dLLM}, as shown in Figure~\ref{fig:q2-cache}.

Both Prefix Cache and Dual Cache substantially improve throughput for all three methods, while accuracy changes remain small. Among the cache-enabled settings, Fast-dLLM achieves the highest throughput, whereas DepCap with Dual Cache provides the best overall balance, reaching 79.3 accuracy with 51.2 TPS and lower NFE than the two cache-enabled AdaBlock variants. These results show that DepCap remains effective after cache integration and combines naturally with existing block-wise cache techniques, which answers Q2.

\subsection{Ablation Study}

\noindent\textbf{To answer Q3.}
To analyze the role of each component in DepCap, we conduct ablation studies on its two core modules. The complete DepCap framework is given by ``DepGA-Block + CAP-Decoding''. We isolate each module through ``Confidence + DepGA-Block'' and ``CAP-Decoding + Fixed Block ($|B_0| = 32$)'', and use ``Vanilla'' as the reference setting where neither module is applied.
This design allows us to separately examine the contribution of each module and the benefit of combining them.

The ablation results, as shown in Table~\ref{tab:reorganized_parallel_decode}, show that both modules contribute to the final gains in different ways. Under standard confidence-based decoding, replacing fixed schedules with DepGA-Block often improves accuracy while keeping throughput competitive, which confirms the value of DepGA-Block. Under the same block size, CAP-Decoding usually achieves the fastest decoding speed with negligible quality degradation, showing that explicit conflict handling allows more aggressive parallel decoding than confidence-only strategy.

When combine the both DepGA-Block and CAP-Decoding, DepCap achieves the best overall trade-off between generation quality and decoding speed, which answers Q3.

\subsection{Hyperparameter Analysis }

\noindent\textbf{To answer Q4.} To study the robustness of DepCap and understand how its behavior changes under different settings, we analyze key hyperparameters. These experiments are intended to show how the speed-quality trade-off changes as the outer DepGA-block and the inner CAP-Decoding become more or less aggressive. Detailed experimental results and analysis are provided in the Appendix~\ref{app:hyper}.

\noindent\textbf{Generation length $L_{\mathrm{gen}}$.}
We study how the behavior of DepCap changes under different generation lengths. The results show that DepCap remains effective across different $L_{\mathrm{gen}}$, although the speed-quality trade-off becomes more challenging as the $L_{\mathrm{gen}}$ grows (i.e., $L_{\mathrm{gen}}=1024$).

\noindent\textbf{Maximum block size $L_\mathrm{max}$.}
We then analyze the effect of the maximum block size in DepGA-Block. As $L_{\max}$ increases, TPS generally improves and NFE decreases. However, overly small settings restrict the parallel generation capability of DLMs, while overly aggressive choices make the generation quality-speed trade-off harder to maintain. 

\noindent\textbf{Uncertainty weight $\lambda$.}
We further analyze the effect of the uncertainty weight $\lambda$. A moderate range of $\lambda$ yields stable performance, while larger values mainly make block expansion more conservative without improving the overall speed-quality trade-off.

\noindent\textbf{Confidence threshold $\tau_{\mathrm{low}}$.}
Next, we study the effect of $\tau_{\text{low}}$ in CAP-Decoding. Lower thresholds favor more aggressive parallel decoding and higher throughput, while larger thresholds improve accuracy but reduce speed.

\noindent\textbf{Conflict threshold $\gamma$.}
Finally, we analyze the effect of the conflict threshold $\gamma$, which determines when two candidates are treated as conflicting. The results show that a looser threshold improves throughput, whereas a more conservative threshold tends to preserve accuracy at the cost of speed.

In a nutshell, the hyperparameter analysis shows that DepCap remains effective across a reasonable range of settings and that its gains do not rely on narrow tuning, which answers Q4.

\section{Conclusion and Discussion}

\noindent\textbf{Conclusion.}
This paper presents DepCap, a training-free framework for efficient block-wise DLM inference.
Rather than determining block boundaries from fixed schedules or current-step local cues, and relying on confidence alone for token-level parallelism, DepCap formulates both as inference-time decisions that should be guided by more suitable signals. 
Specifically, cross-step signals from the last decoded block guide block partitioning, while token conflict signals control within-block parallel decoding, enabling a better quality-speed trade-off.
We also provide an information-theoretic analysis of the dependency score by relating cumulative last-block influence to block partitioning decisions. 
Empirically, DepCap consistently improves decoding efficiency with negligible quality degradation, and remains compatible with existing block-wise cache techniques.

\noindent\textbf{Discussion.}
The current work focuses on block-wise DLM inference. A related and equally important direction is sliding-window diffusion decoding~\citep{DCD,DSB,ElasticCache}, where how to adapt the window size by incorporating information from the historical multi-step denoising process is itself an interesting problem. In addition, the present theory only provides an information-theoretic analysis of the dependency score and does not yet establish a strict bound or stronger formal guarantee. Strengthening this analysis, and extending the underlying ideas beyond the block-wise setting, are important directions for our future work.

\bibliographystyle{plainnat} 
\bibliography{mycite}


\appendix
\newpage
\input{appendix.tex}



\end{document}

%% file: appendix.tex
\section{Detailed Theoretical Analysis}
\label{app:Analysis}

\subsection{Detailed Derivation}

We fix the decoded context $c$ before the latest block and write that block as a random variable $B \sim p(B \mid c)$. The next candidate future block is $Z_{1:L} = (Z_1,\dots,Z_L)$. For each future position $k$, the analysis compares the predictive distributions before and after decoding the last block:
\begin{equation}
  p_k^{\text{prev}}(\cdot) = p(Z_k=\cdot \mid c),
  \qquad
  p_k^{\text{curr}}(\cdot) = p(Z_k=\cdot \mid c, B).
\end{equation}
The corresponding last-block influence for a realized block is
\begin{equation}
  I_k^{\mathrm{sample}}(b)
  =
  \KL\!\left(
    p(Z_k \mid c, b)
    \,\middle\|\,
    p(Z_k \mid c)
  \right).
\end{equation}

\subsection{Expected Influence Versus Sample Influence}

The expected influence term used in the analysis is
\begin{equation}
  I_k^{\mathrm{exp}}
  =
  \I(Z_k; B \mid c).
\end{equation}
By the definition of conditional mutual information,
\begin{equation}
  \I(Z_k; B \mid c)
  =
  \E_{B \sim p(\cdot \mid c)}\!\left[
    \KL\!\left(
      p(Z_k \mid c, B)
      \,\middle\|\,
      p(Z_k \mid c)
    \right)
  \right].
\end{equation}
Hence
\begin{equation}
  I_k^{\mathrm{exp}}
  =
  \E\!\left[I_k^{\mathrm{sample}}(B)\right].
\end{equation}
This is why it is consistent to analyze DepCap through conditional mutual information while implementing it with a KL score computed for one realized block.

The block-level quantity of interest is the cumulative information that the last decoded block provides to the next $L$ positions:
\begin{equation}
  \mathcal{I}(L)
  =
  \I(Z_{1:L}; B \mid c).
\end{equation}
By the chain rule,
\begin{equation}
  \I(Z_{1:L}; B \mid c)
  =
  \sum_{k=1}^{L}
  \I(Z_k; B \mid c, Z_{<k}).
\end{equation}
Writing each term as
\begin{equation}
  \I(Z_k; B \mid c, Z_{<k})
  =
  \I(Z_k; B \mid c)
  -
  \Bigl(
    \I(Z_k; B \mid c)
    -
    \I(Z_k; B \mid c, Z_{<k})
  \Bigr),
\end{equation}
and summing over $k$ yields the exact identity
\begin{equation}
  \mathcal{I}(L)
  =
  \sum_{k=1}^{L} I_k^{\mathrm{exp}} - \varepsilon(L),
  \label{eq:appendix-correction}
\end{equation}
where
\begin{equation}
  \varepsilon(L)
  =
  \sum_{k=1}^{L}
  \Bigl(
    \I(Z_k; B \mid c)
    -
    \I(Z_k; B \mid c, Z_{<k})
  \Bigr).
\end{equation}
This is the overlap correction caused by shared information from the last decoded block across future positions.

Under Assumption~\ref{ass:local-overlap}, the overlap term remains moderate for the short local blocks used by DepCap, so
\begin{equation}
  \mathcal{I}(L)
  \approx
  \sum_{k=1}^{L} I_k^{\mathrm{exp}}.
\end{equation}
In this sense, the per-position last-block influence terms can be viewed as local contributions to the cumulative support supplied by the last decoded block.

Influence alone does not determine whether the next block should keep expanding, because support from the last decoded block must be compared against how uncertain the future positions remain. DepCap therefore also considers
\begin{equation}
  H_k = \Entropy(Z_k \mid c, B).
\end{equation}
A natural design objective is
\begin{equation}
  \mathcal{I}(L)
  \geq
  \lambda \sum_{k=1}^{L} \E[H_k].
\end{equation}
Substituting Equation~\eqref{eq:appendix-correction} yields
\begin{equation}
  \sum_{k=1}^{L} I_k^{\mathrm{exp}}
  -
  \lambda \sum_{k=1}^{L} \E[H_k]
  \geq
  \varepsilon(L).
\end{equation}
If $|\varepsilon(L)|$ remains moderate over the local candidate window, the sign pattern is mainly governed by the per-position terms. Dropping the correction then gives the simpler approximation
\begin{equation}
  \sum_{k=1}^{L}
  \bigl(I_k^{\mathrm{exp}} - \lambda \E[H_k]\bigr)
  \geq 0.
\end{equation}
The online dependency score
\begin{equation}
  S_k = I_k^{\mathrm{sample}} - \lambda H_k
\end{equation}
is the sample version of the summand above, so it measures whether the support from the last decoded block still outweighs uncertainty at position $k$.

\subsection{Implementation Approximation}

The derivation above uses the theoretical distributions $p_k^{\text{prev}}$ and $p_k^{\text{curr}}$. In the actual implementation, DepCap instead compares the predictive logits immediately before and after the latest block is decoded while keeping the rest of the inference state fixed. We therefore view the implemented score as a computationally tractable approximation to the theoretical last-block influence, rather than as an exact evaluation of the corresponding distributions.

This approximation is most faithful when the dominant change between the two inference states is precisely the effect of decoding the latest block. This is also the setting in which DepCap is intended to operate: short local blocks whose effect on nearby future positions can be estimated reliably from the change in the model's predictive distributions.

If last-block influence decays as the decoding process moves away from the frontier while uncertainty remains non-negligible, then $S_k$ naturally transitions from positive to near-zero and then negative. The first negative crossing is therefore the first point at which further expansion receives too little support from the last decoded block relative to its uncertainty. This gives a local stopping rule whose cumulative interpretation is to stop once support no longer dominates uncertainty.

\section{Detailed Hyperparameter Analysis}
\label{app:hyper}

\subsection{Generation Length}

\begin{table}[h]
  \centering
  \caption{Effect of generation length $L_{\mathrm{gen}}$ on DepCap on LLaDA-8B-Instruct over GSM8K.}
  \label{tab:appendix-lgen}
  \begin{tabular}{c|ccc}
    \toprule
    $L_{\mathrm{gen}}$ & Acc $\Uparrow$ & TPS $\Uparrow$ & NFE $\Downarrow$ \\
    \midrule
    128  & 74.5 & 19.1 & 48.4 \\
    256  & 78.8 & 21.4 & 74.8 \\
    512  & 79.8 & 15.5 & 99.2 \\
    1024 & 77.0 & 8.7  & 120.4 \\
    \bottomrule
  \end{tabular}
\end{table}

\noindent As shown in Table~\ref{tab:appendix-lgen}, increasing $L_{\mathrm{gen}}$ makes the decoding problem progressively harder. From 128 to 512 tokens, accuracy improves moderately, which suggests that DepCap can still exploit parallel decoding effectively as the generation horizon grows. At the same time, TPS decreases steadily and NFE increases substantially, indicating that longer sequences require more denoising iterations and are harder to accelerate efficiently. When $L_{\mathrm{gen}}$ further increases to 1024, accuracy also begins to drop, showing that very long generations make the quality-speed trade-off more difficult to maintain.

\subsection{Uncertainty Weight}

\begin{table}[h]
  \centering
  \caption{Effect of the uncertainty weight $\lambda$ on DepCap on LLaDA-8B-Instruct over GSM8K.}
  \label{tab:appendix-lambda}
  \begin{tabular}{c|ccc}
    \toprule
    $\lambda$ & Acc $\Uparrow$ & TPS $\Uparrow$ & NFE $\Downarrow$ \\
    \midrule
    0.6 & 78.8 & 20.6 & 75.1 \\
    0.8 & 78.9 & 20.7 & 74.9 \\
    1.0 & 79.0 & 20.8 & 74.8 \\
    1.2 & 78.8 & 21.4 & 74.8 \\
    1.4 & 78.8 & 20.7 & 74.8 \\
    1.6 & 78.6 & 20.8 & 74.9 \\
    \bottomrule
  \end{tabular}
\end{table}

\noindent As shown in Table~\ref{tab:appendix-lambda}, the uncertainty weight $\lambda$ mainly controls how conservatively DepGA-Block expands the next block. Across the tested range, the variation in accuracy, TPS, and NFE is relatively small, which suggests that DepGA-Block is not overly sensitive to the exact choice of $\lambda$. Moderate values around $1.0$--$1.2$ provide the strongest overall balance, while larger values make block expansion more conservative without yielding better overall speed-quality trade-offs.

\subsection{Maximum Block Size}

\begin{table}[h]
  \centering
  \caption{Effect of the maximum block size $L_{\max}$ on DepCap on LLaDA-8B-Instruct over GSM8K.}
  \label{tab:appendix-lmax}
  \begin{tabular}{c|ccc}
    \toprule
    $L_{\max}$ & Acc $\Uparrow$ & TPS $\Uparrow$ & NFE $\Downarrow$ \\
    \midrule
    16  & 79.6 & 17.6 & 90.8 \\
    32  & 80.1 & 19.1 & 83.6 \\
    64  & 78.5 & 20.7 & 77.5 \\
    128 & 78.8 & 21.4 & 74.8 \\
    256 & 79.5 & 21.5 & 74.8 \\
    \bottomrule
  \end{tabular}
\end{table}

\noindent As shown in Table~\ref{tab:appendix-lmax}, larger $L_{\max}$ generally leads to higher TPS and lower NFE, since DepGA-Block is allowed to form longer candidate blocks whenever the last-block influence remains informative. However, the accuracy trend is not monotonic. Very small settings, such as $L_{\max}=16$, clearly restrict the parallel generation capability of DLMs, while overly aggressive settings favor speed but make the quality-speed trade-off harder to maintain. These results suggest that $L_{\max}$ controls the ceiling of parallelism in DepGA-Block and should be chosen large enough to avoid unnecessary restriction, but not so large that quality becomes harder to preserve.

\subsection{Candidate Threshold}

\begin{table}[h]
  \centering
  \caption{Effect of the candidate threshold $\tau_{\text{low}}$ on DepCap on LLaDA-8B-Instruct over GSM8K.}
  \label{tab:appendix-taulow}
  \begin{tabular}{c|ccc}
    \toprule
    $\tau_{\text{low}}$ & Acc $\Uparrow$ & TPS $\Uparrow$ & NFE $\Downarrow$ \\
    \midrule
    0.65 & 78.4 & 24.8 & 61.7 \\
    0.70 & 78.5 & 24.3 & 65.5 \\
    0.75 & 78.7 & 18.4 & 69.8 \\
    0.80 & 78.8 & 21.4 & 74.8 \\
    0.85 & 80.3 & 19.0 & 80.3 \\
    \bottomrule
  \end{tabular}
\end{table}

\noindent As shown in Table~\ref{tab:appendix-taulow}, the candidate threshold $\tau_{\text{low}}$ directly controls how broadly CAP-Decoding expands the candidate pool before conflict filtering. Smaller thresholds make decoding more aggressive, leading to much higher TPS and lower NFE, but they also reduce accuracy. Larger thresholds are more selective and improve accuracy, but this comes at the cost of slower decoding. This behavior is consistent with the role of $\tau_{\text{low}}$: it determines the trade-off between exploring a larger parallel candidate set and preserving generation quality.

\subsection{Conflict Threshold}

\begin{table}[h]
  \centering
  \caption{Effect of the conflict threshold $\gamma$ on DepCap on LLaDA-8B-Instruct over GSM8K.}
  \label{tab:appendix-gamma}
  \begin{tabular}{c|ccc}
    \toprule
    $\gamma$ & Acc $\Uparrow$ & TPS $\Uparrow$ & NFE $\Downarrow$ \\
    \midrule
    -8.0  & 78.9 & 23.8 & 67.2 \\
    -12.0 & 78.4 & 22.8 & 70.1 \\
    -16.0 & 78.8 & 21.4 & 74.8 \\
    -20.0 & 79.5 & 19.9 & 80.6 \\
    -24.0 & 79.5 & 17.8 & 86.3 \\
    \bottomrule
  \end{tabular}
\end{table}

\noindent As shown in Table~\ref{tab:appendix-gamma}, the conflict threshold $\gamma$ determines how strictly CAP-Decoding removes conflicting candidates inside the current block. A looser threshold improves throughput, with the best TPS achieved at $\gamma=-8.0$, while more conservative thresholds gradually reduce speed and increase NFE. Accuracy remains relatively stable across the tested range, which suggests that $\gamma$ mainly controls how aggressively CAP-Decoding exploits within-block parallelism after the candidate pool has already been formed.